\documentclass{article}

 \usepackage[nonatbib, final]{nips_2016}

\usepackage{graphics} 
\usepackage{epsfig} 
\usepackage{mathptmx} 
\usepackage{times} 
\usepackage{amsmath} 
\usepackage{amssymb}  
\usepackage{fixltx2e}
\usepackage{caption}
\usepackage{subcaption}
\usepackage[colorlinks]{hyperref}

\usepackage[utf8]{inputenc} 
\usepackage[T1]{fontenc}    
\usepackage{hyperref}       
\usepackage{url}            
\usepackage{booktabs}       
\usepackage{amsfonts}       
\usepackage{nicefrac}       
\usepackage{microtype}      

\title{End-to-End Deep Reinforcement Learning for Lane Keeping Assist}

\author{Ahmad El Sallab$^{1}$, Mohammed Abdou$^{1}$, Etienne Perot$^{2}$ and Senthil Yogamani$^{3}$
\thanks{$^{1}$Ahmad El Sallab is a chief engineer and Mohammed Abdou is a researcher at Valeo,
        Cairo, Egypt
        {\tt\small ahmad.el-sallab@valeo.com}}%
\thanks{$^{2}$Etienne Perot is a research engineer at Valeo Paris, 
        France
        {\tt\small etienne.perot@valeo.com}}%
\thanks{$^{3}$Senthil Yogamani is a technical lead at Valeo Vision Systems, 
        Ireland
        {\tt\small senthil.yogamani@valeo.com}}%
}

\begin{document}

\maketitle

\begin{abstract}
Reinforcement learning is considered to be a strong AI paradigm which can be used to teach machines through interaction with the environment and learning from their mistakes, but it has not yet been successfully used for automotive applications. There has recently been a revival of interest in the topic, however, driven by the ability of deep learning algorithms to learn good representations of the environment. Motivated by Google DeepMind's successful demonstrations of learning for games from Breakout to Go, we will propose different methods for autonomous driving using deep reinforcement learning. This is of particular interest as it is difficult to pose autonomous driving as a supervised learning problem as it has a strong interaction with the environment including other vehicles, pedestrians and roadworks. As this is a relatively new area of research for autonomous driving, we will formulate two main categories of algorithms: 1) Discrete actions category, and 2) Continuous actions category. For the discrete actions category, we will deal with Deep Q-Network Algorithm (DQN) while for the continuous actions category, we will deal with Deep Deterministic Actor Critic Algorithm (DDAC). In addition to that, We will also discover the performance of these two categories on an open source car simulator for Racing called (TORCS) which stands for The Open Racing car Simulator. Our simulation results demonstrate learning of autonomous maneuvering in a scenario of complex road curvatures and simple interaction with other vehicles. Finally, we explain the effect of some restricted conditions, put on the car during the learning phase, on the convergence time for finishing its learning phase. 
\end{abstract}

\section{INTRODUCTION}

The Reinforcement Learning (RL) framework \cite{c24} \cite{c29} has been used in control tasks for some time. The mixture of RL with DL was noted to be one of the most promising approaches to achieve human-level control in \cite{c13}. In \cite{c17} and 
\cite{c16}, professional or even superhuman performance was achieved on Atari games using the Deep Q Networks (DQN) model. In DQN, RL is responsible for the planning part of the model, while DL is responsible for the representation learning part. More recently, RNNs have been integrated into the model to account for partially observable scenarios \cite{c7}. Driving task is a critical task that needs high level of skills, attention, and experience from the driver. This means that reaching autonomous driving is an extreme challenge especially for the machine intelligence.

Autonomous driving task can be categorized into three main parts: \\
1) Exploration: It is responsible for discovering the surrounding environment objects like: traffic signs and lights, pedestrians detection, lane detection, etc. Nowadays, these tasks are relatively easier because of the success of Deep Learning algorithms and the performance has reached the level of human capabilities for the object detection and recognition \cite{c12} \cite{c3}. As an example of potential sensors, camera installed in front of the car (at the dashboard) can be used to capture images and then be fed to a CNN. Although camera images are very high dimensional, the useful information needed for the task of autonomous driving is of much lower dimension.
Deep learning models are able to learn complex feature representations from raw input data, omitting the need for hand-crafted features \cite{c21} \cite{c3} \cite{c11}. In this regard, Convolutional Neural Networks (CNNs) are probably the most successful deep learning model, and have formed the basis of every winning entry on the ImageNet challenge since AlexNet \cite{c12}. This success has been replicated in lane and vehicle detection for autonomous driving \cite{c9}.
In simulation, TORCS supported us with these sensors thanks to the Simulated Car Racing (SCR)  plug-in that provides us with the sensor models and readings.

2) Model Solution: It is considered as the most important division for our task because it is the variable part for any system to solve a problem. As the exploration division could be the same for many experiments, but the algorithm used to create the solution is the different. We depend on the exploration division in order to generate or create the efficient model for our aim. As a result, We will follow two main and basic categories: 1) Discrete Actions Algorithms, and 2) Continuous Actions Algorithms. 

3) Testing: It is considered as the phase that validates the success of the algorithm used to create the model. We can not only test the created model but also compare it with the performance of various algorithms.

In this paper, the main contributions are: 1) proposing different methods for end-end autonomous driving model that takes raw sensor inputs and outputs driving actions, 2) presenting a survey of the recent advances of deep reinforcement learning, and 3) following the previous system (Exploration, Model Solution and testing) in order to: a) compare between the performance of the two main categories: discrete actions, and continuous actions algorithms; b) compare between the performance of the same algorithm under some restricted conditions applied while running these experiments; and c) discover the effect of these restricted conditions on the convergence time of learning. 

\section{DEEP REINFORCEMENT LEARNING}

Depending on the problem domain, the space of possible actions may be discrete or continuous, a difference which has a profound effect on the choice of algorithms to be applied. In this section we will discuss two algorithms: one which operates on discrete actions (DQN) and one which operates on continuous actions (DDAC). Our deep reinforcement learning framework paper \cite{c34} provides more details on how the mentioned methods below are combined together. For a comprehensive overview of reinforcement learning, please refer to the second edition of Rich Sutton's textbook \cite{c25}. We provide a brief overview of essential topics.

\subsection{Deep Q Networks (DQN)} 

When the states are discrete, the Q-function can be easily formulated as a table. This formulation becomes harder when the number of states increases, and even impossible when the states are continuous. In such case, the Q-function is formulated as a parameterized function of the states, actions;$Q(s,a,w)$. The solution then lies in finding the best setting of the parameter $w$ . Using this formulation, it is possible to approximate the Q-function using a Deep Neural Network (DNN). The objective of this DNN shall be to minimize the Mean Square Error (MSE) of the Q-values as follows:

\begin{gather*}
l(w) = E[(r + \gamma \arg\max_{a'} Q_{t} (s',a',w) - Q_{t} (s,a,w))^2] \\
J(w) = \max_{w} l(w)
\end{gather*}
  
This objective function is differentiable end-end in with respect to its parameters, i.e. $\frac{\partial l(w)}{\partial w}$  exists. Thus, the optimization problem can be easily solved using Gradient based methods (Stochastic Gradient Descent (SGD), Conjugate Gradients (CG),etc). The algorithm is called Deep Q-Networks (DQN) \cite{c17}\cite{c16}.
	
\subsection{Deep Deterministic Actor Critic (DDAC)} 

The DQN algorithm is suitable for continuous states cases, but the action selection still requires the action values to be discrete. Several algorithms were suggested for continuous actions cases, where two functions are learned: 1) the actor; which provides the policy mapping from a state to action, and 2) the critic; which evaluates (criticizes) the value of the action taken in the given state. The critic function is the same as the Q-function. The algorithms to learn both function follow the policy gradient methods \cite{c26}. Under the framework of deep learning, both functions can be learned through two neural networks; $Q(s,a,w)$  and $\pi(s,u)$ , since the whole objective is still differentiable w.r.t. the weights of the Q-function and the policy. Hence, the gradient of the Q-function (the critic) is obtained as in DQN: $\frac{\partial l(w)}{\partial w}$ , while the gradient of the policy function (the actor) is obtained using the chain rule as follows: 
\begin{gather*}
\frac{\partial J}{\partial u} = \frac{\partial Q}{\partial a} \vert_{a=\pi(s,u)} \frac{\partial \pi(s,u)}{\partial u}
\end{gather*}

 \subsection{Deep Recurrent Reinforcement Learning}

 The Q-learning algorithms are based on the Markov assumption of the MDP. In situations where the full observability of the environment is not available, this assumption is not valid anymore. Partially observable MDP (POMDP) arises in different scenarios in autonomous driving, like the occlusion of objects during tracking, mapping and localization. POMDPs are tackled using information integration over time, where the true state of the environment is not directly revealed from single observation, but gradually form over multiple observations at different time stamps. The recurrent neural networks (RNN) present themselves as a natural framework to tackle POMDPs. In \cite{c19} RNN was successfully applied  for the task of end to end multi-object tracking. Moreover, LSTMs \cite{c8} are integrated to the DQNs to form the Deep Recurrent Q Networks (DRQN) in \cite{c7}. The application of DRQN in \cite{c7} to Atari games does not show the full power of the model, since the MDP assumption is usually enough for Atari games, hence the authors try a variant of the Pong game; Flickering Pong, they show the advantage of adding recurrence.

\subsection{Deep Attention Reinforcement Learning}

 In the DQN model, the spatial features are extracted via a CNN, which learns the features from data. Those features are not all contributing equally to the final optimization objective. Similar to the recognition process in human beings, only a limited amount of information is needed to perform the recognition tasks, and not all the high dimensional sensory data. Attention models \cite{c32} are trying to follow the same concept, where only part of the CNN extracted features are used in the classification task. This part is learned in parallel to the original learning process. In \cite{c15}, a separate network called “glimpse network” is trained to deploy the kernel operation at certain parts of the image. The motion of the kernel is learned via the REINFORCE algorithm \cite{c31} to learn the best sequence of motions of the kernel over the input image. The result is a motion that resembles the reading process in human beings in case of feeding images of digits for example. The idea is exploited in \cite{c23} to improve the DQN and DRQN models, by including a filter after the CNN features extractor, resulting in the DARQN model. The DARQN was shown to attend and highlight parts of the Atari games that are of special importance. In the “Sequest” game, the agent was able to focus on the oxygen level of the submarine, while in “Breakout” game; the agent was able to track the ball position \cite{c31}.

\subsection{Apprenticeship learning}

The Reinforcement learning algorithms described so far follow the concept of “episodic” learning, or “learning from delayed rewards” \cite{c29}. In this setting, the rewards function is assumed to be known to the algorithm. However, in some cases, instead of having clear mapping of a state to reward function, we have a demonstrated expert behavior to the agent. The goal of the agent in that case is to decipher the intention of the expert, and decode the structure of the reward function. This is referred to as Inverse Reinforcement learning \cite{c1}. The reward function is encoded as a linear combination of “features” functions that maps the state according to some features of interest. For example, for the driving task, one feature function could be: “how far is the car from the lanes”. 
Another approach is described in \cite{c5}, where the demonstrated expert behavior is utilized in a supervised model, where the actions taken by the expert together with the states are considered as the training examples to a CNN.

\section{DRL SYSTEM FOR LANE KEEPING ASSIST}

In this section, we will deal with a DRL system for Lane Keeping Assist. All of these methods will depend on the input come from the environment in which we could construct the input states, and the outputs are the driving actions for the car autonomously.   

\subsection{Data Collection and Environment Exploration}

The first stage in our system is Data Collection and Environment Exploration. Exploration for the environment is achieved by sensors like: Camera, LIDAR, RADAR, etc. We depend on these sensor readings for our algorithms because autonomous driving requires the integration of all information from these sensors (Sensor Fusion). The requirement for the DNN to extract relevant features from raw sensor input makes sensor fusion a natural task in the course of the learning process.
The state of the surrounding objects is usually not directly observed, but rather deduced by the algorithm through a set of sensors readings. Fusing such sensor information is mandatory to make use of them. Sensor fusion is a wide area of research by itself. The car state includes its position, orientation, velocity and acceleration. In the case of autonomous driving, the surrounding environment state needs to be encoded as well. The environment state may include objects, their location, orientation, motion vectors, dimensions, etc. Traditional autonomous driving algorithms would make use of these state vectors to plan the motion path in such an environment.
On the other hand, an end-end deep learning system would use a different encoding of the state. For example in \cite{c17}, the states are just the snapshots of the game, which include by default a lot of implicit information. Such information is not explicitly given to the algorithm, but rather deduced through the DNN (more specifically a CNN), in the form of “features”. 

\subsection{Model Creation Algorithms}
In this section, we will provide details for the algorithms that are used for our autonomous driving. We have two main categories for our algorithms: 1) Discrete Actions Algorithms, and 2) Continuous Actions Algorithms. This enables us to apply some examples on both of the two categories and then compare between their performance.  

\subsubsection{Discrete Actions Algorithms}

\subsubsection*{Q-Learning Algorithm}
We started the reinforcement learning with Q-Learning Algorithm\cite{c10}\cite{c30}. This algorithm depends on creating a Q-table which is normally considered as a map for the environment states, due to the fused sensor readings, for the discretized actions taken. Due to the continuous environment and the discrete algorithm, we used a function approximator called: tile coding. We have predefined tiles for the controlled actions, this means that the controlled actions are approximated to the nearest one. This algorithm has been empirically successful, but it has many drawbacks such as the high confidence level in one estimator only for the optimal future Q-value while taking the future into our consideration in the Q-learning equation.

\subsubsection*{Deep Q-Network (DQN) Algorithm}
The normal progress for the reinforcement learning is to go to the Deep Reinforcement learning. The Q-function is modified to be not only function in the states and the actions taken but also in the weights of a DNN to estimate the optimal future Q-value. This modification solves the drawback of the Q-learning algorithm, but it still deals with discretized actions following the function approximator.    

\subsubsection{Continuous Actions Algorithms}
\subsubsection*{Deep Deterministic Actor Critic (DDAC) Algorithm}
It is an On-Policy Deep Learning Algorithm in which the policy is independent on value function. It depends on the Actor-critic algorithm, in which we have two networks: 1) Actor Network: DNN responsible for taking actions based on the states, and 2) Critic Network: DNN responsible for criticizes the value of the action taken in the given state. As we know that the main difference between DQN and DDAC is the action taken is discrete or continuous respectively. However, both of them follow the same training procedure with a Q-function on the top having the same objective function. In case of continuous actions, the DDAC network can be used. The same error is back propagated through the network structure to obtain the gradients at every network layer.
This algorithm solves the great drawback of the Q-Learning, and DQN algorithms

\section{RESULTS AND DISCUSSION}

\begin{figure}[htbp]
    \centering
    \includegraphics[scale=0.45]{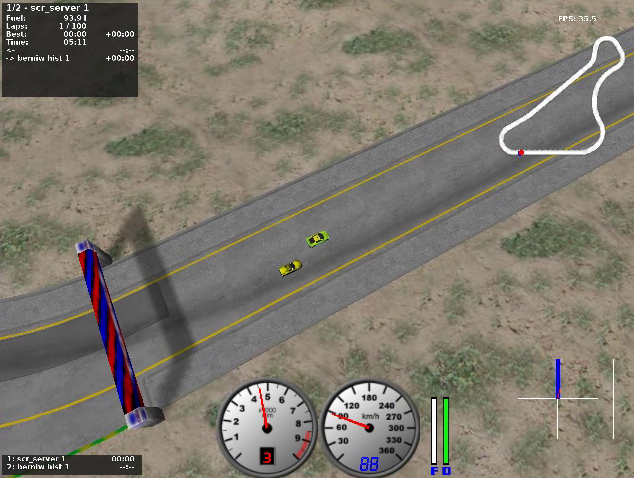}
    \caption{TORCS screen-shot of DRL based lane keeping}
    \label{fig:label4}
  \end{figure}

In this section, we discuss how we tested Lane keeping assist function using TORCS. The Open Racing Car Simulator (TORCS) is used in order to benefit from its plug-in which is Simulated Car Racing (SCR). TORCS-SCR provide us with the car model, various tracks, graphics and physics engine beside the car control parameters like: steering angle, velocity, acceleration, brakes. In addition to that, it provides full information for any real car like: Car position, velocity, fuel level, RPM... etc. The provided information is not only for the car but also for the surrounding environment for the track and the opponents on the same track as well. In terms of simulation setup, the input to the network is the \textit{trackPos} sensor input, which provides the position of the track borders, in addition to the car speed in the x-position. The output are the steering, the gear, the acceleration and the brake values. The network is trained end-end following the same objective of the DQN. Figure \ref{fig:label4} is a sample screenshot of the lane keeping scenario.

In order to formulate the application as a classification problem, the actions (steer, gear, brake, acceleration) are tiled and discretized. In another setup, the actions are taken as continuous values following the policy gradient method of the DDAC. The results show successful lane keeping function in both cases. However, in case of DQN, removing the replay memory trick (Q-learning) helps to have faster convergence and better performance. In addition, the tile coding of actions in case of DQN makes the steering actions more abrupt. On the other hand, using DDAC continuous policy helps smoothing the actions and provides better performance.

We tested both of Q-learning Algorithm and DDAC Algorithm on the same track whose shape as in Fig.(1). This track is characterized by containing both of straight and curved parts which is good for testing both algorithms well. For the straight part of the track, we found that both of Q-learning Algorithm and DDAC Algorithm have approximate performance for surviving on the track as in Fig.(2). For curved part of the track, there is a great difference in the performance of both algorithms: for Q-learning algorithm as in Fig.(2a), we noticed that the car takes Discrete actions and there is some time between the current action and the next action, this is obvious from the car path in the figure, while for DDPG algorithm as in Fig.(2b), we noticed that the car takes continuous actions so there is a smooth curve as shown in the path. We can conclude that both of the two algorithms are reasonable for the straight part of the track, on the other side, the Q-learning is not reasonable for the curved part of the track and the DDAC algorithm has an excellent performance on both straight and curved part of the track.

\begin{figure}
	\begin{subfigure}{.5\textwidth}
  	\centering
  	\includegraphics[width=.6\linewidth]{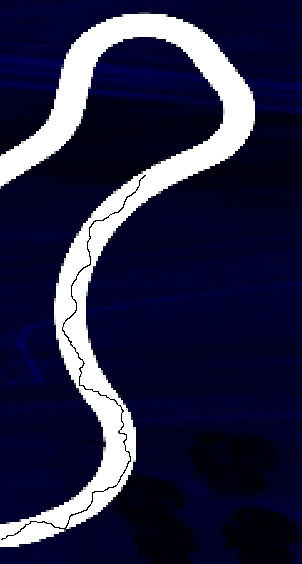}
    \includegraphics[width=.2\linewidth]{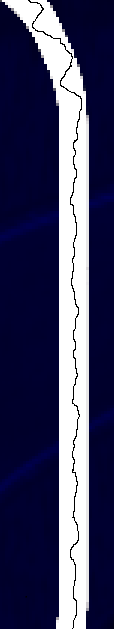}
  	\caption{Q-Learning Performance}
  	\label{fig:sfig1}
	\end{subfigure}
	\begin{subfigure}{.5\textwidth}
  	\centering
  	\includegraphics[width=.6\linewidth]{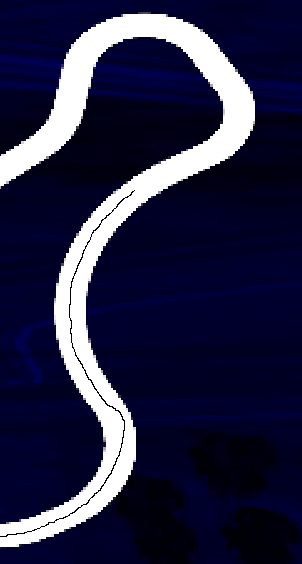}
    \includegraphics[width=.2\linewidth]{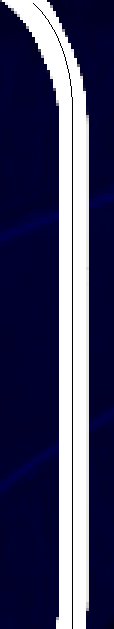}
  	\caption{DDAC Performance}
  	\label{fig:sfig2}
	\end{subfigure}
	\caption{Q-Learning vs. DDAC Performances on the same Track}
	\label{fig:fig}
\end{figure}

\subsection{TORCS Termination Conditions}

Termination criterion is important as it indicates whether the training phase has converged to the expected goal. We have many termination conditions used especially on the DDAC algorithm: (1) No Termination condition, (2) Out of Track, (3) Stuck, and (4) Out of Track with Stuck. 

\subsubsection{No Termination Condition}

The car can move in any part of the track without any restricted termination condition, but it takes only negative Rewards if it does wrong actions like: get out of the track or be in a stuck mode. The only termination condition is when the Car becomes in horizontal way on the track; we can reach to this mode when the car tries to take sharp curve on the track. If the car gets out of the track, it takes a negative Reward, but we don’t terminate TORCS. We allow the car to explore the track by itself and learn from its action on the track, so there is no other restricted termination condition. This termination condition is considered as the basic one and it is applied on other experiments, so normally it exists and no need to talk about it again.

\subsubsection{Out of Track}

When the car gets out of the track, it takes a high negative reward, and then we terminate TORCS and re-launch it again, this means that we start a new episode. This means that during the learning phase, we prevent the car from getting out of the track, so it is expected that during the testing phase, the car doesn’t get out of the track.

\subsubsection{Stuck}

We give the car around 100 time steps, which is equivalent to the initial motion of the car (initial acceleration), then we put a termination condition which is if the car stalls or stuck, we terminate TORCS and re-launch it again; this means that we start a new episode. The stall or the stuck condition depends on the vertical speed of the car (speedX) in which the stuck concept is if the speed reaches to (5 km/h), this is considered as a tuning parameter to prevent the car from reaching low speeds. This also means during the learning phase, we prevent the car from approaching to the low speeds, so it is expected that during the testing phase, the car will not approach from these low speeds. This termination condition is very useful for the aim of Racing.

\subsubsection{Out of Track with Stuck}

This experiment has a hybrid termination conditions which are getting out of the track and Stuck termination conditions. We applied both of these restricted termination conditions. This means that we started a new episode if one of them happens or occurs.
 
\subsection{Convergence Time vs. Termination Conditions}

\subsubsection{No Termination Condition}

It is expected that this experiment converges at low number of episodes. This is due to the absence of some restrictions of the termination conditions. Lesser the number of terminations, faster the convergence time. One episode takes long time to be terminated; this enables the car to complete one lap as fast as possible, to explore the whole track more and more as fast as possible. On the other side, we are afraid of settling on a local minimum point for the neural networks, due to the long time learning on the same track, that may lead the car stuck either between the track boundaries or out of the track. This gives us the permission to add some restricted termination conditions like: out of track termination condition and Stuck termination condition.

\subsubsection{Out of Track}

It is expected that this experiment converges at moderate number of episodes greater than the number of episodes in the Basic Experiment. This is because of starting new episode in case of the car gets out of the track, so the car needs many episodes to explore the whole track and to complete one lap for that track.

\subsubsection{Stuck}

It is expected that this experiment converges at moderate number of episodes greater than the number of episodes in the Basic Experiment. This is because of starting new episode in case of the car stuck, so the car needs many episodes to explore the whole track and to complete one lap for that track.
Stuck Termination condition can help both the basic experiment and out of track experiment, so it is most commonly used with the basic experiment in order to avoid settling in a local minimum either the car found between the track boundaries or the car settles out of the track. This enables us also to add another condition which is the out of track termination condition.

\subsubsection{Out of Track with Stuck}

It is expected this experiment converges at high number of episodes greater than all other experiments. This is due to the presence of two restricted conditions, so both of them lead to start new episode either car stuck or car gets out of the track. We restricted the car from exploring the track by itself because we enforced it not to stuck and not get out of the track.

\subsection{Experimental Results for the Termination Conditions}

In order to measure the level of learning for the same track for the whole experiments, we defined that if the car completed 10 laps per episode means that the car learned that track well. The next graph defines the measure of learning for the four experiments.

\begin{figure}[htbp]
	\centering
	\includegraphics[scale=0.5]{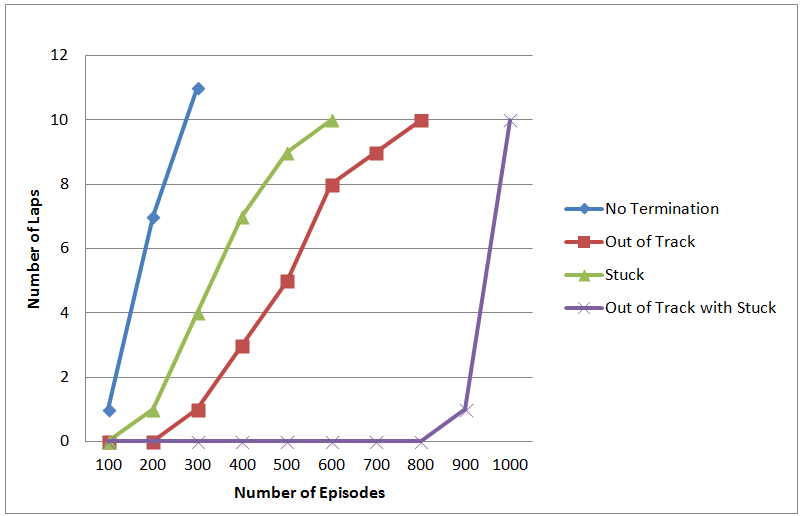}
	\caption{Termination Conditions effect on Convergence Time}
	\label{fig:label6}
\end{figure}

It is obvious that when there was no termination conditions, the car learned faster than using some restricted termination condition. As we see that Learning with no termination conditions converges faster than Learning with Stuck termination condition faster than out of track termination condition faster than combining two termination conditions of out of track and Stuck. This is because whatever the action taken, if it is a good action, so it takes a positive reward, otherwise it takes a negative reward. In addition to that, the one episode takes a long time to be terminated, so the car could complete one lap from the first episode which helps the car to explore the full track faster. On the other hand, in case of having some restricted termination conditions, TORCS is forced to be terminated which means starting a new episode, so it is a trade-off we don't want the car do this action again, but we lost the advantage of fast convergence of no termination conditions.       

\section{CONCLUSION}

In this paper, we introduced DRL system for lane keeping assist depending on different categories for the used algorithms. These categories differ from each other in the type of the actions taken by the car whether: discrete or continuous. After that we compared between two specific algorithms: Q-learning whose actions are discrete, and DDAC whose actions are continuous from the performance point of view and the smoothness actions on the same track. Then, we introduced a new field of research which is studying the effect of the restricted conditions (termination conditions) on the convergence time of learning for the same algorithm. We concluded that the more we put termination conditions, the slower convergence time to learn. 




\section*{APPENDIX}

Sample DRL training and demo sequences are provided as supplementary material for the review process. Please visit the following youtube links for \href{https://youtu.be/hktC8sGURJQ}{DRL training using DQN}, \href{https://youtu.be/OtuKpWew6UI}{DRL training using DDAC} and \href{https://youtu.be/RxIkdKGtzTE}{DRL lane keeping using regression neural network}. \\

Entering the URLs explicitly in case the hyperlinks are suppressed.\\
DRL training using DQN - https://youtu.be/hktC8sGURJQ \\
DRL training using DDAC - https://youtu.be/OtuKpWew6UI \\
DRL lane keeping using regression neural network - https://youtu.be/RxIkdKGtzTE \\

\section*{ACKNOWLEDGMENT}

The authors would like to thank their employer for the opportunity to work on fundamental research. Thanks to B Ravi Kiran (INRIA France) and Catherine Enright (Valeo) for reviewing the paper and providing feedback. Special thanks to a colleague Matthew Carrigan for detailed review, proof-reading and editing.



\end{document}